\documentclass[wcp]{jmlr}

\usepackage{longtable}

\usepackage{booktabs}

\usepackage{lineno}
\nolinenumbers
\usepackage{amsmath}
\usepackage{amssymb}

\makeatletter
\let\Ginclude@graphics\@org@Ginclude@graphics 
\makeatother

\jmlrvolume{}
\jmlryear{2025}
\jmlrworkshop{ACML 2025}

\title[CON-QA]{CON-QA: Privacy-Preserving QA using cloud LLMs in Contract Domain}

\author{%
  \Name{Ajeet Kumar Singh}$^{1,2}$ \Email{ajeetkumar.singh@infoorigin.com}\\
  \Name{Rajsabi Surya}$^{1}$ \Email{rajsabi.surya@infoorigin.com}\\
  \Name{Anurag Tripathi}$^{1}$ \Email{anurag.tripathi@infoorigin.com}\\
  \Name{Santanu Choudhury}$^{2}$ \Email{santanuc@ee.iitd.ac.in}\\
  \Name{Sudhir Bisane}$^{1}$ \Email{sudhirb@infoorigin.com}\\
  \addr $^{1}$ Info Origin, INC \\
  \addr $^{2}$ Indian Institute of Technology, New Delhi (IITD)
}

\editors{}

\begin{document}

\maketitle

\begin{abstract}
As enterprises increasingly integrate cloud-based large language models (LLMs) such as ChatGPT and Gemini into their legal document workflows, protecting sensitive contractual information—including Personally Identifiable Information (PII) and commercially sensitive clauses—has emerged as a critical challenge. In this work, we propose CON-QA, a hybrid privacy-preserving framework designed specifically for secure question answering over enterprise contracts, effectively combining local and cloud-hosted LLMs. The CON-QA framework operates through three stages: (i) semantic query decomposition and query-aware document chunk retrieval using a locally deployed LLM analysis, (ii) anonymization of detected sensitive entities via a structured one-to-many mapping scheme, ensuring semantic coherence while preventing cross-session entity inference attacks, and (iii) anonymized response generation by a cloud-based LLM, with accurate reconstruction of the original answer locally using a session-consistent many-to-one reverse mapping. To rigorously evaluate CON-QA, we introduce CUAD-QA, a corpus of 85k question-answer pairs generated over 510 real-world CUAD contract documents, encompassing simple, complex, and summarization-style queries. Empirical evaluations, complemented by detailed human assessments, confirm that CON-QA effectively maintains both privacy and utility, preserves answer quality, maintains fidelity to legal clause semantics, and significantly mitigates privacy risks, demonstrating its practical suitability for secure, enterprise-level contract documents.
\end{abstract}

\section{Introduction}
The widespread adoption of cloud-based large language models (LLMs), such as OpenAI’s ChatGPT~\cite{achiam2023gpt} and Google’s Gemini \cite{team2023gemini}, has significantly transformed enterprise AI applications. These models offer sophisticated natural language processing capabilities, proving invaluable for diverse business tasks including information retrieval, summarization, contract analysis, and question answering (QA) \cite{cheng2024remoterag}.
However, the integration of cloud-based LLMs into enterprise workflows raises critical concerns about data privacy and security. Specifically, transmitting sensitive business information, such as Personally Identifiable Information (PII) and confidential clauses within legal contracts, to external APIs poses significant risks \cite{demir2025legalguardian} \cite{chen2023hide}. These include potential data leakage, regulatory non-compliance, and unauthorized data access \cite{abualhaija2022automated}. Such risks are especially pertinent in regulated industries, where the protection of contractual information is mandated by frameworks like GDPR \cite{j2024data} and HIPAA \cite{tschider2024new}. Consequently, there is a pressing need for robust privacy-preserving approaches that can effectively leverage cloud LLMs without compromising sensitive enterprise information. While many large firms possess the resources to develop and deploy proprietary LLM models tailored to their specific needs, others lack access to such advanced and costly infrastructure. This disparity underscores the need for lightweight, accessible frameworks designed specifically for domain-specific contract QA.
To address these challenges, we introduce CON-QA, a hybrid privacy-preserving question-answering framework specifically designed for secure enterprise legal contracts. The CON-QA framework systematically combines locally (inhouse) and cloud-based LLMs in a three-stage pipeline: (a) semantic query decomposition and retrieval of relevant contract chunks through local LLM analysis, (b) detection and anonymization of sensitive entities via a carefully designed one-to-many mapping scheme that maintains semantic integrity while preventing entity linkage across different user sessions, and (c) generation of anonymized answers using the cloud-based LLM, followed by accurate local reconstruction of responses through a deterministic many-to-one reverse mapping strategy. By discarding entity mappings after each session, our design significantly mitigates the risk of re-identification and cross-session inference attacks, which further enhances privacy protection.
Unlike conventional anonymization approaches, which can degrade semantic coherence, CON-QA retains the fidelity and legal integrity of contractual content throughout the anonymization and reconstruction process. To evaluate our framework comprehensively, we curate CUAD-QA, a dataset comprising approximately 85k QA pairs generated over 510 contracts of the CUAD corpus~\cite{hendrycks2021cuad}. The dataset realistically simulates enterprise-level user queries, spanning simple, complex, and summarization tasks commonly encountered in practical contract query scenarios. Empirical evaluations and rigorous human assessments conducted on CUAD-QA demonstrate that CON-QA consistently achieves strong privacy preservation without sacrificing answer quality or the semantic nuances critical to legal contract interpretation.

Our main contributions in this paper are summarized as follows:
\begin{itemize}
\item We propose CON-QA, a session-based, hybrid privacy-preserving QA framework for contract analysis. It employs a one-to-many anonymization scheme that generates diverse, semantically aligned replacements for sensitive PII in both queries and retrieved chunks, enabling secure and utility-preserving answer generation.

\item We construct the CUAD-QA dataset, which simulates realistic enterprise-level queries on contractual documents, encompassing approximately 85k question-answer pairs generated on the CUAD dataset \cite{hendrycks2021cuad}.

\item We conduct extensive empirical evaluations, including rigorous human assessments, validating that CON-QA robustly protects sensitive information while preserving the accuracy, semantic fidelity, and legal coherence required in practical enterprise contract query.
\end{itemize}




\section{Related Works}

\subsection{NLP in the Legal Domain}

 From the past few years, the success of NLP in specialized domains has been noteworthy. The legal domain, specifically, has been one of the most researched topics when it comes to NLP and automation. This is because the legal domain comprises a mountain of sensitive text documents that require in-depth analysis. \cite{hendrycks2021cuad} introduced the Contract Understanding Atticus Dataset (CUAD), which identifies 41 categories of clauses that are crucial in corporate transactions such as mergers and acquisitions. The authors also benchmarked the dataset by leveraging BERT, ALBERT, RoBERTa, and DeBERTa for the objective of predicting which substrings of a contract relate to each label category.
 Introducing LLMs to the legal sector, \cite{colombo2024saullm} developed two language models, namely SaulLM-54B and SaulLM-141B, based on a Mixture of Experts architecture. The Mixtral model was fine-tuned on several legal datasets, including the FreeLaw subset and the MultiLegal Pile. The models outperform existing open-source LLMs, like GPT-4 and LLaMA 3, on legal benchmarks such as LegalBench-Instruct, demonstrating the effectiveness of scaling up both model and corpus size for domain specialization.
 With the substantial growth of NLP in the legal domain, the topic of privacy and security becomes essential. \cite{demir2025legalguardian} proposed LegalGuardian to preserve privacy in the prompts used by lawyers by masking sensitive Personally Identifiable Information (PII) with fake entities. They used a synthetic data set to reflect the workflow of an immigration lawyer. GLiNER \cite{zaratiana2024gliner} Multi PII-v1 and Qwen were utilized to identify PII and replace them with fake entities before sending them to an external LLM. Dictionaries were maintained for masking and unmasking, with masking performed at the PII Masking Layer and unmasking achieved at the Secure Prompting Layer.
Building on this line of research, we focus on the enterprise contract domain, where privacy-preserving question answering demands even more robust anonymization and recovery mechanisms due to the complexity and sensitivity of contractual data.

\subsection{Privacy Preservation in LLMs}
\cite{chen2023hide} present a lightweight framework, Hide and Seek (HaS), for prompt privacy protection. For training, synthetic data is generated with a news summarization corpus as the base dataset. The data is used to train two Small Language Models for anonymization and deanonymization purposes. On the user side, the hide model is tasked with hiding private entities. The seeker model receives the anonymized input from the hide model, the original query, and the output of the cloud LLM as input, while it outputs the de-anonymized result.
On similar grounds, \cite{hou2025general} propose a pseudoanonymization framework for QA , classification, machine translation, summarization, and text generation tasks. The authors suggest identifying private entities via NER, prompt-based, or seq2seq methods and generating replacement entities with similar characteristics, either via random sampling or prompt-based techniques. Finally, the replacement candidates are substituted by using direct replacement, prompt-based replacement, or replacement through text generation.
\cite{carpentier2024preempting} addresses the problem of resource wastage in the context of privacy-preserving LLMs. They state that applying privacy mechanisms to prompts (like Differential Privacy) can lead to the generation of useless answers from the LLM. To mitigate this, a novel utility assessor is proposed, comprising an SLM to check the quality and utility of masked prompts. This helps in assessing the necessary amount of privacy to introduce based on the utility assessor's evaluation, thereby preventing LLMs from generating low-utility responses and saving computational resources.
\cite{frikha2024incognitext} propose IncogniText, which aims to enhance privacy against attribute inference attacks without compromising the meaning or semantics of a query by misleading an adversary. IncogniText uses an adversarial model to mimic a potential attacker by identifying the author's attribute value. An anonymization model is then leveraged to misdirect the attacker by replacing or anonymizing the attribute value. Experiments were conducted using Meta LLaMA \cite{lotfian2025performance}, Phi, and Qwen models \cite{bai2023qwen} as adversary, anonymization, and attacker modules.
The problem of privacy leakage in cloud-based RAG systems was identified by \cite{cheng2024remoterag}. To address this, they propose Remote-RAG, in which the user introduces random noise to their query (Embedding Perturbation), which satisfies an (n, $\epsilon$)-DistanceDP privacy guarantee before sending it to the cloud. The cloud then retrieves the top k' documents for the perturbed query. The authors prove that these k' documents contain the top k documents corresponding to the original query as well, ensuring that accuracy is not compromised, while utilizing partially homomorphic encryption. we observe that differential privacy–based mechanisms, such as DistanceDP, struggle with semantically rich, domain-specific scenarios like enterprise contract QA. Specifically, the added perturbation can obscure legally significant clauses, disrupt clause-level retrieval, or degrade interpretability in downstream legal reasoning. To address these limitations, we propose CON-QA—a privacy-preserving framework tailored for contractual QA over cloud-based LLMs. Con-QA maintains query intent and semantic fidelity through structured anonymization and deterministic deanonymization. Building on advancements in anonymization, adversarial obfuscation, and utility-aware privacy trade-offs, it introduces session-specific one-to-many surrogate mapping, ensures high surrogate diversity, and employs a deterministic reverse mapping scheme. This end-to-end design yields a robust, black-box-compatible solution for secure and accurate enterprise contract question answering.

\section{Sensitive ContractQA Dataset (CUAD-QA)}
\label{sec:dataset}
 To enable the evaluation of privacy-preserving mechanisms in enterprise legal settings, we construct a large-scale dataset, CUAD-QA, generated over the CUAD \cite{hendrycks2021cuad}(Contract Understanding Atticus Dataset) corpus. Unlike general-purpose QA datasets focused on open-domain or encyclopedic content, CUAD-QA targets the unique demands of contractual question answering, where queries often reference commercially sensitive entities, clause semantics, and jurisdiction-specific obligations.
\paragraph{ Motivation and Privacy Relevance}
 Enterprise contracts frequently contain personal, organizational, and financial identifiers, such as: Company names and subsidiaries, Named individuals and signatories, Jurisdictions, governing law, and court reference, Proprietary technologies, payment thresholds, and exclusivity terms. When such documents are queried using cloud-based LLMs, they pose significant privacy and compliance risks, especially under legal frameworks like GDPR \cite{tschider2024new}, HIPAA \cite{j2024data}, and corporate confidentiality agreements. The semantic richness and fine-grained clause structure of CUAD make it an ideal benchmark for studying these risks and testing anonymization strategies.
 
\subsection{Dataset Construction and Purpose}

To model realistic enterprise-level interactions with contractual documents, we construct a large-scale question answering dataset based on the CUAD (Contract Understanding Atticus Dataset). We employ ChatGPT-4o-mini \cite{achiam2023gpt} to generate natural, clause-grounded questions and corresponding reference answers by prompting it with individual pages from 510 full-length contract documents. This process yields a total of 85k QA pairs, averaging approximately 10 QA pairs per contract page, three query types commonly encountered in contract review question presented in Table\ref{tab:my-Table_1} (i){Simple questions}, which involve direct lookups from a single clause. (ii){Complex questions}, which require multi-clause reasoning and interpretation, and (iii) \textit{Summarization questions}, which demand abstractive synthesis of multiple clauses or obligations. anonymization, semantic fidelity testing, and expert human evaluation of privacy-preserving QA. This dataset underpins the development and evaluation of the CON-QA framework, enabling systematic exploration of the privacy--utility tradeoffs in cloud-based contract question answering, including deanonymization accuracy and answer quality recovery across anonymized answer. examples of each question type along with their reference answers are illustrated in Table~\ref{tab:my-Table_1}.


\begin{table}[ht]
\centering
\caption{Sample Questions and Reference Answers by Query Type in the CON-QA Dataset}
\label{tab:my-Table_1}
\begin{tabular}{p{3cm} p{6cm} p{6cm}}
\toprule
\textbf{Query Type} & \textbf{Sample Question} & \textbf{Reference Answer} \\
\midrule
Simple & What is the effective date of this agreement? & January 1, 2023 \\
\addlinespace

Complex & Does the exclusivity clause apply to all product categories in North America? & Yes, the exclusivity applies to all categories listed in Exhibit B for the North American market. \\
\addlinespace
Summarization & Summarize the buyer’s obligations in the event of early termination. & The buyer must provide 30 days' written notice, settle outstanding payments, and return any leased equipment. \\
\bottomrule
\end{tabular}
\end{table}

\section{Proposed Approach}
{Enterprise applications often involve processing sensitive contractual information, posing a significant risk when interacting with cloud LLMs such as ChatGPT \cite{achiam2023gpt}, Qwen \cite{bai2023qwen}. These models, while offering state-of-the-art (SOTA) capabilities, are typically accessed via remote APIs, making them opaque and untraceable to the end users. In such black-box environments, any direct exposure of confidential contract clauses, named entities, or financial terms risks irreversible privacy leakage. To address this, we propose CON-QA, a privacy-preserving question answering tailored for enterprise contract documents, a hybrid framework that integrates local LLM reasoning capabilities with cloud-based LLM  for answer generation while ensuring strong privacy guarantees. CON-QA is designed to protect sensitive information details during interactions with cloud-based large language models (LLMs). The session-based, hybrid privacy-preserving QA framework consists of three major components: (a)Query Analysis and RAG-Based Document Chunk Retrieval, (b)PII detection and one-to-many entity mapping set generation for anonymization, (c)Cloud Answer Generation and deterministic Deanonymization in original terms using a many-to-one reverse entity mapping scheme. The detailed explanation is provided in each subsection, along with an overview and the mechanism of our framework, as illustrated in Figure~\ref{fig:fig_label_0}.
\begin{figure*}[htb!]
  \centering
  \centerline{\includegraphics[width = 0.9\linewidth,height = 4in]{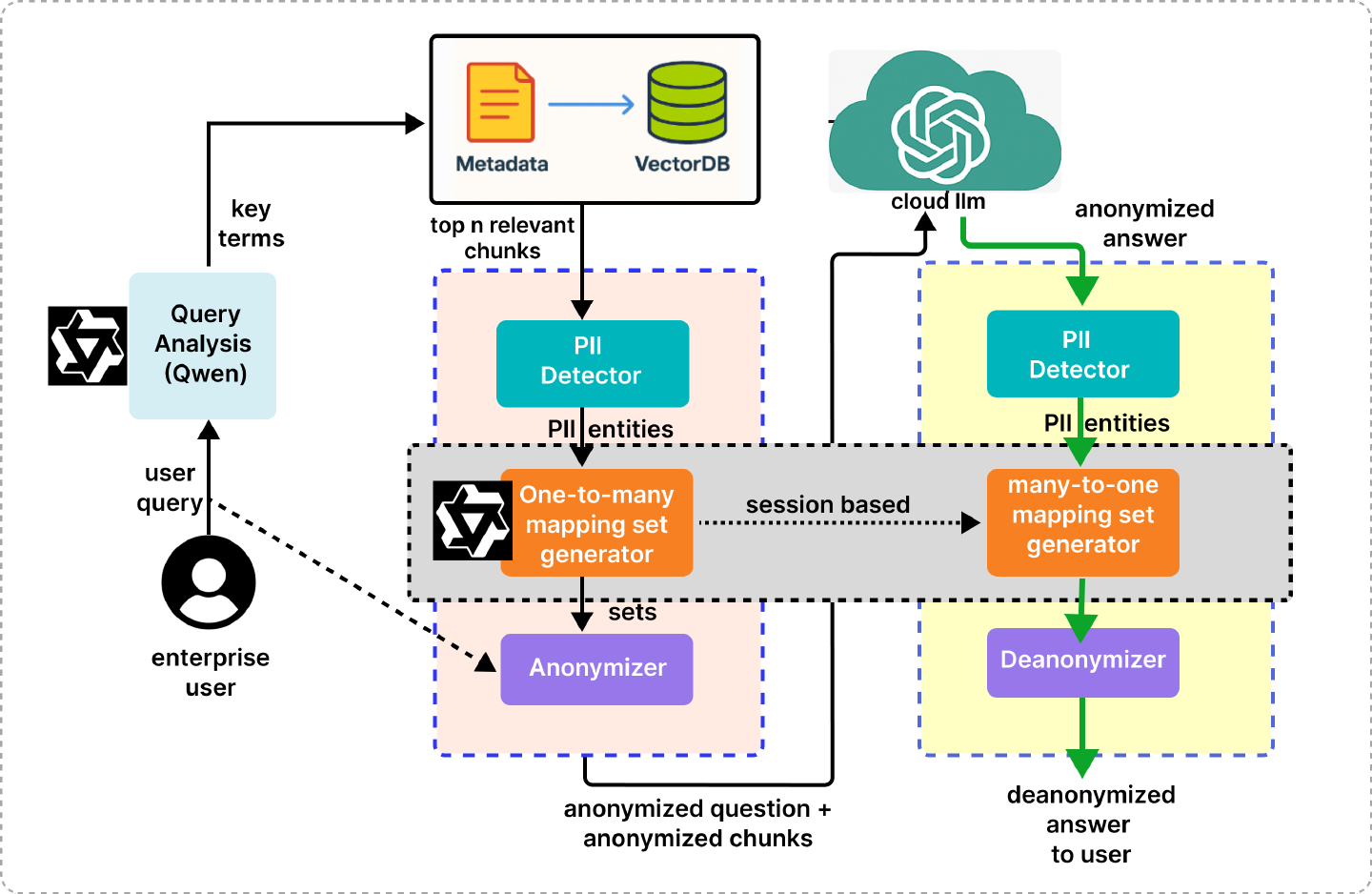}}
  \caption{CON-QA Pipeline Architecture: A privacy-preserving QA framework with session-based anonymization and deterministic deanonymization.}
\label{fig:fig_label_0}
\end{figure*} 
\subsection{Query Analysis and RAG-Based Document Chunk Retrieval}
Given a user query posed against a local repository of enterprise contracts, our framework first employs a locally hosted large language model (Qwen-2.5-14B) to perform query decomposition and semantic scope identification. This step is designed not for final inference, but to carry out early-stage linguistic parsing and contract-aware retrieval in a controlled, privacy-sensitive environment.We adopt a retrieval-augmented generation (RAG) strategy to extract the most relevant document chunks from the local contract corpus. This enables the framework to isolate only the semantically pertinent fragments required for answering the query.By offloading these upstream tasks to a locally deployed LLM, CON-QA achieves a privacy-conscious balance: sensitive contractual content is semantically filtered and processed in-house before any interaction with external models. This hybrid design ensures that the cloud LLM, used downstream for answer generation, receives only a tightly scoped and privacy-preserving representation of the original input, significantly reducing the risk of inadvertent information leakage.

Let the user query be \( X \). The objective is to extract a structured set of semantic components relevant to contract understanding. Let these components be denoted as:

\[
A = \{ a_i \mid a_i \in X,\, 1 \leq i \leq m \}
\]

where each \( a_i \) corresponds to a specific semantic category such as document identifiers, named entities, clause types, temporal expressions, or query intent. Formally, we define:

\[
A = \{ \text{doc\_ids},\, \text{parties},\, \text{metadata\_fields},\, \text{text\_search\_terms},\, \text{query\_type},\, \text{dates} \}
\]

To extract these components, we use a locally installed Qwen2.5-14B-Instruct-1M model. The model is prompted with schema-guided instructions to map the natural language input \( X \) into its structured representation \( A' \), where:

\[
A' = \{ a'_i \mid a'_i \in A,\, \text{extracted via Qwen},\, 1 \leq i \leq m \}
\]

Post-processing heuristics are applied to refine the output.

We construct a document-level metadata index for the CUAD corpus by analyzing all contract documents using the same Qwen2.5 model. This index is stored as \texttt{metadata.json}, where each entry links a document to its semantic attributes.

Given the extracted fields \( A' \), we search this metadata to identify the relevant document IDs:

\[
D = \{ d_k \mid d_k \in \texttt{metadata.json},\, \text{matches}(d_k, A') \}
\]

These document IDs \( D \) are used to retrieve the corresponding documents from the corpus. For each \( d_k \in D \), a semantic similarity search is conducted over its associated text chunks using a vector database. The most relevant chunks are selected for downstream response generation.


\subsection{Anonymyzation: privacy preserving using one-to-many entity mapping set generation }
In this anonymization stage, private entities detected in the user query and the retrieved document chunks are extracted using a named entity recognition model, GlinNER \cite{zaratiana2024gliner}. For each identified entity, a one-to-many mapping set is generated consisting of multiple substitute entities that are semantically and linguistically consistent with the original. This mapping ensures that the anonymized content retains the original intent and contextual meaning, enabling effective downstream processing while preserving privacy. Here, we break down the framework into its key elements and describe their individual functions.

\subsubsection{Sensitive Information Detection in Query and Chunks}

Let the user query be denoted by \( Q \), and the set of top-\( k \) retrieved document chunks be represented as \( D = \{d_1, d_2, \ldots, d_k\} \). To identify sensitive information, we define a named entity recognition (NER) operator \( \mathcal{N}(\cdot) \), which maps a given input text to a set of extracted sensitive entities.

The set of sensitive entities extracted from the query \( Q \) is defined as:
\[
\mathcal{E}_Q = \{e_i^Q \mid e_i^Q \in \mathcal{N}(Q),\ 1 \leq i \leq n_Q \}
\]

Similarly, for each retrieved chunk \( d_j \in D \), the extracted sensitive entities are:
\[
\mathcal{E}_{d_j} = \{e_i^{d_j} \mid e_i^{d_j} \in \mathcal{N}(d_j),\ 1 \leq i \leq n_{d_j} \}
\]

The complete set of sensitive entities detected across the query and all retrieved chunks is given by:
\[
\mathcal{E}_{\text{total}} = \mathcal{E}_Q \cup \left( \bigcup_{j=1}^{k} \mathcal{E}_{d_j} \right)
\]

Here, \( \mathcal{E}_{\text{total}} \) represents the union of all potentially sensitive named entities that will be subjected to anonymization. The operator \( \mathcal{N}(\cdot) \) is implemented using the GlinNER model \cite{zaratiana2024gliner}.
In the context of contract documents, the NER operator  \( \mathcal{N}(\cdot) \)  is designed to extract typical sensitive entity categories such as person names, company names, geographic locations (e.g., cities, states, street,country), and other legally significant identifiers. These entities often carry privacy risk and are systematically extracted prior to anonymization. The NER function is implemented using the GlinNER model \cite{zaratiana2024gliner} (Liu et al., 2023), which supports high-accuracy detection of both simple and contextually complex named entities in legal text.

\subsubsection{One-to-Many Entity Mapping Set Generation}

Let the set of sensitive entities extracted from the user query and retrieved document chunks be denoted as
\[
\mathcal{E}_{\text{total}} = \{ e_1, e_2, \ldots, e_N \}.
\]
For each entity \( e_i \in \mathcal{E}_{\text{total}} \), we aim to generate a corresponding anonymization set
\[
\mathcal{S}_i = \{ s_{i1}, s_{i2}, \ldots, s_{iK} \}
\]
containing \( K \) surrogate entities that are semantically similar to \( e_i \), i.e.,
\[
s_{ij} \approx_{\text{sem}} e_i \quad \forall j \in \{1, \ldots, K\}.
\]

This one-to-many mapping is achieved via in-house prompting of the Qwen model \cite{yang2025qwen2}, which receives each \( e_i \) as input and returns a disjoint, semantically coherent, and internally diverse set of \( K \) candidates. The objective is to preserve the intent and functional meaning of the original entities while ensuring resistance against reverse inference.

\paragraph{Formal Mapping Structure.} The resulting one-to-many mapping across all sensitive entities is defined as:
\[
\mathcal{M} = \left\{ (e_i, \mathcal{S}_i) \,\middle|\, e_i \in \mathcal{E}_{\text{total}},\ |\mathcal{S}_i| = K,\ \text{sim}(e_i, s_{ij}) \geq \theta,\ \text{dist}(s_{im}, s_{in}) \geq \delta \right\}.
\]

This formulation allows principled construction of anonymization candidates that (i) preserve semantic meaning, (ii) ensure categorical separation, and (iii) introduce internal variation—balancing fidelity, privacy, and adversarial robustness. The resulting mappings serve as the foundation for randomized anonymization and subsequent deanonymization procedures in the pipeline.

\paragraph{Session-Specific Randomized Replacement.}

After generating the mapping set \( \mathcal{M} = \{(e_i, \mathcal{S}_i)\} \), a session-specific randomized replacement strategy is employed. For each original entity \( e_i \), a single anonymized substitute \( \tilde{e}_i \) is uniformly sampled from its corresponding set \( \mathcal{S}_i \).That is, one candidate is selected at random and used to replace the original entity in both the user query and the retrieved chunks.These replacements are performed independently for each session. To maintain privacy and eliminate the possibility of cross-session linkage, the anonymization sets \( \mathcal{S}_i \) are discarded once the session ends. As a result, the same original input entity \( e_i \) can produce different anonymized versions \( \tilde{e}_i \) across different sessions.

This session-specific, non-deterministic mapping approach strengthens privacy guarantees by preventing deterministic substitution patterns and reducing the risk of reverse inference or entity re-identification over time.
\[
\tilde{e}_i = \phi^{(t)}(e_i) \sim \text{Uniform}(\mathcal{S}_i), \quad \forall e_i \in \mathcal{E}_{\text{total}}, \quad \text{where } \phi^{(t)} \text{ is discarded after session } t.
\]
\noindent
This denotes that each original entity \( e_i \) is replaced by a randomly selected surrogate \( \tilde{e}_i \) from its corresponding set \( \mathcal{S}_i \), using a session-specific mapping \( \phi^{(t)} \) that is ephemeral and deleted after the session ends.


\subsection{Deanonymization: Using many-to-one reverese entity mapping}

Once the anonymized user query and the associated anonymized, semantically relevant document chunks are processed by a cloud-based large language model, GPT-4o-mini \cite{achiam2023gpt}, we obtain a response \( \tilde{A} \) that is formulated using anonymized surrogate entities \( \tilde{e}_i \in \mathcal{S}_i \).
Let the anonymized answer be:
\[
\tilde{A} = f_{\text{LLM}}(\tilde{Q}, \tilde{D}),
\]
where \( \tilde{Q} \) is the anonymized query, \( \tilde{D} \) is the anonymized document chunks input, and \( f_{\text{LLM}} \) denotes the response function computed by the GPT-4o-mini model.

\paragraph{Many-to-One Reverse Mapping.}

To recover the answer in original terms, both the anonymized question and the corresponding anonymized answer are subjected to a second pass of PII detection using Gliner \cite{zaratiana2024gliner}. Using the stored one-to-many entity mapping \( \mathcal{M} = \{(e_i, \mathcal{S}_i)\} \) generated during the anonymization phase, we define the inverse mapping function \( \phi^{-1} \) as:

\[
\phi^{-1}(\tilde{e}_i) = e_i, \quad \text{where } \tilde{e}_i \in \mathcal{S}_i.
\]

This many-to-one reverse mapping is applied over the anonymized answer \( \tilde{A} \), resulting in the final de-anonymized output:

\[
A = \phi^{-1}(\tilde{A}),
\]

where all surrogate entities are replaced with their original counterparts. The mapping \( \phi^{-1} \) is constructed solely within the session and is discarded immediately after the de-anonymization step:

\[
\phi^{-1} \rightarrow \varnothing \quad \text{after completion of session}.
\]

This session-specific mapping protocol ensures that no persistent links exist between anonymized and original entities across sessions, thereby upholding strong privacy guarantees and preventing reverse inference or cross-session leakage.
This setup enables the cloud-based language model to operate exclusively on privacy-preserving inputs, ensuring that sensitive information is never exposed to third-party systems. By preserving the structural and semantic fidelity of both the query and its associated document chunks—even in the presence of simple, complex, or highly summarized queries from the CUAD dataset—the model remains capable of producing meaningful anonymized outputs. The subsequent many-to-one reverse mapping scheme effectively recovers the original entities with high fidelity. Overall, this approach strikes a strong balance between privacy and utility, offering robust protection against entity leakage, re-identification, and cross-session linkage. Different stages of the query and the relevant chunks are shown in ~\ref{fig:egWalkthrough}.

\begin{figure*}[t]
\centering
\includegraphics[width=17cm, height = 12cm]{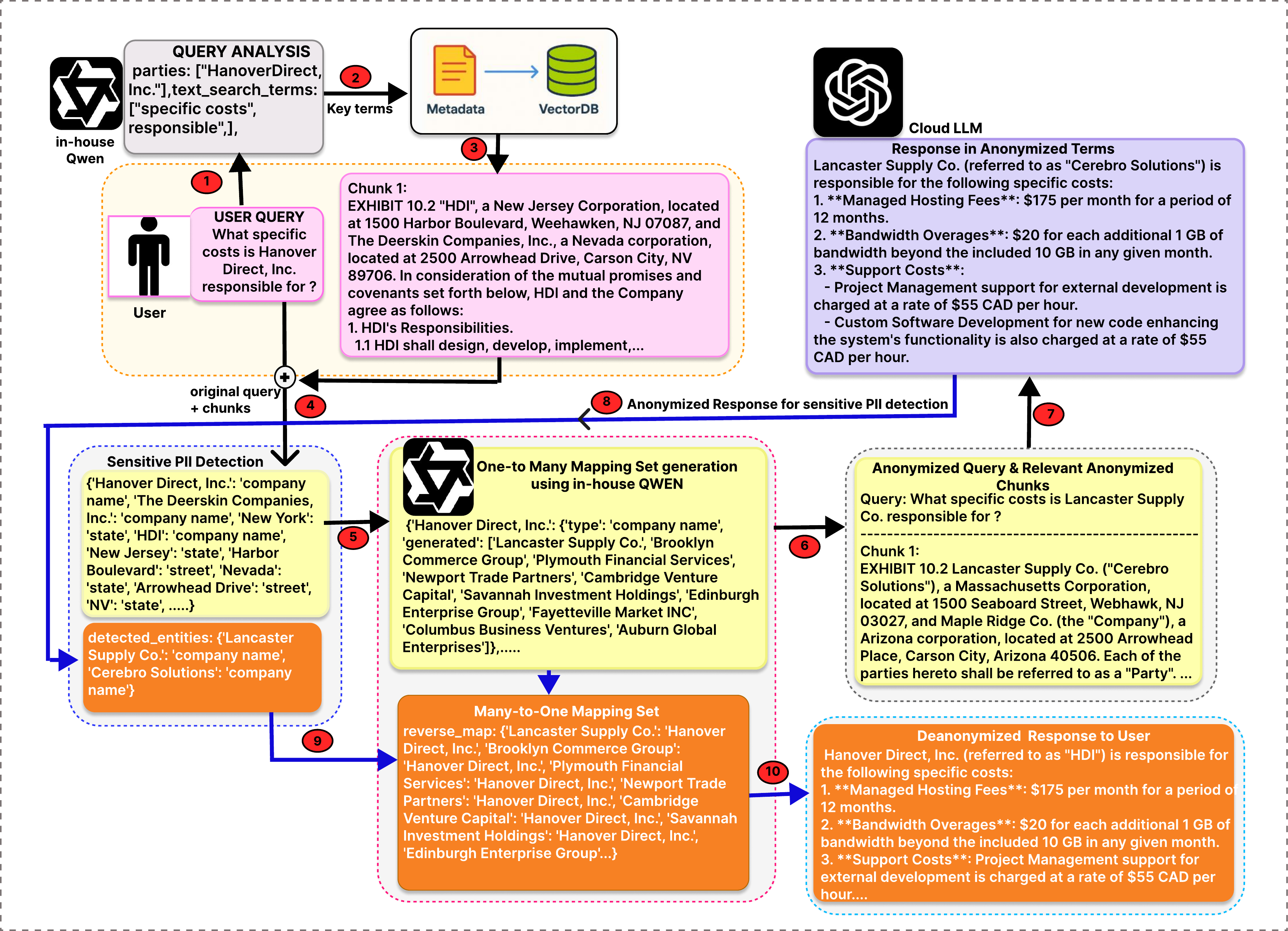}

\caption{ End-to-End example walkthrough of the CON-QA Framework: Demonstration of anonymization and deterministic recovery for a contract query over sensitive enterprise data.}

\label{fig:egWalkthrough}
\end{figure*}

\section{Experiment}
In this work, we evaluate the performance of the proposed CON-QA framework using 1000 QA sampled out from the CUAD-QA dataset section \ref{sec:dataset}. The dataset includes a diverse set of query simulating realistic scenarios in enterprise-level contract document QA.
Our CON-QA pipeline is specifically designed for querying enterprise contract documents in a privacy-preserving environment. The pipeline leverages cloud-based Large Language Models (LLMs) while ensuring that sensitive personal and confidential corporate information is not exposed to these black-box cloud models. The anonymization and deanonymization mechanisms within our pipeline ensure entity-level confidentiality throughout the query-answering process.We evaluate our framework comprehensively across four critical metrics.In addition to these evaluations, we also present comparative analyses of the anonymization and deanonymization components of CON-QA with existing HaS framework \cite{chen2023hide}. Specifically, we compare our anonymization strategy against the Hide model and our deanonymization recovery mechanism against the Seek model from the HaS framework. These comparative results are presented in separate subsections to highlight the relative strengths and limitations of each approach.
\begin{itemize}
    \item \textbf{Private Entity Restoration Accuracy}: This metric assesses the effectiveness of our anonymization-deanonymization pipeline by measuring the accuracy of restoring private entities. Specifically, we compare answers obtained from our complete pipeline, which includes anonymization and subsequent deanonymization, with answers generated without applying these privacy-preserving steps. This comparison explicitly quantifies the impact of our privacy mechanism on the integrity and accuracy of the final answers.
    \item \textbf{Response Relevancy}: Leveraging the SOTA Ragas metrics, the response relevancy metric quantifies how pertinent the generated responses are to the user's input query. 
    \item \textbf{Answer Correctness}: The answer correctness comprises two essential components: semantic similarity and factual accuracy. Semantic similarity measures how closely the generated answers align in meaning with the ground-truth answers. Factual similarity evaluates the precision and correctness of factual information within the generated responses relative to the ground truth.

    \item \textbf{Faithfulness}: Faithfulness assesses the factual consistency between the recovered responses and the relevant retrieved documents. This metric ensures that the responses faithfully represent the factual content available in the original document context without introducing inconsistencies or hallucinated content.

    \end{itemize}
\begin{table}[h]
\centering
\caption{Evaluation of CON-QA on Key Performance Metrics}
\label{tab:conqa_metrics}
\begin{tabular}{lc}
\toprule
\textbf{Metric} & \textbf{Score} \\
\midrule
Private Entity Restoration Accuracy & 0.9880 \\
Response Relevancy                  & 0.9778 \\
Answer Correctness                  & 0.8810 \\
Faithfulness                        & 0.9886 \\
\bottomrule
\end{tabular}
\end{table}

\paragraph{CON-QA Key Performance Metrics Analysis}The evaluation results presented in Table \ref{tab:conqa_metrics} demonstrate robust performance of the CON-QA framework across all assessed metrics. Notably, the private entity restoration accuracy of 0.9880 highlights the high reliability of our anonymization-deanonymization process, confirming minimal loss of information integrity in the privacy-preserving mechanisms.
Response relevancy achieved a score of 0.9778, underscoring the pipeline's capability to consistently provide pertinent responses aligned closely with user queries.The Answer Correctness score of 0.8810 reflects a modest decline, largely attributable to summarization-style queries where reference answers are brief, while CON-QA’s recovered answers are more contextually enriched and elaborative. This semantic expansion, while accurate, affects strict string-based matching. To address this, we conduct expert human evaluation detailed in a subsequent subsection—across four nuanced correctness-reasoning categories to better capture answer fidelity beyond surface-level similarity.
A high faithfulness score of 0.9886 underscores CON-QA’s strong factual alignment with source context, affirming its reliability. Overall, the framework achieves a robust balance between privacy preservation and accurate, and context-rich contract QA.

\paragraph{Implementation details}:We implement the CON-QA framework using a hybrid setup comprising a locally hosted LLM (Qwen-2.5-14B-Instruct-1M) and a cloud-based LLM (GPT-4o-mini). The Qwen model is employed for CUAD metadata extraction, query analysis, and generating one-to-many surrogate mappings. GPT-4o-mini is used for CUAD-QA generation and anonymized answer retrieval. All experiments are conducted on a dual-GPU setup with NVIDIA RTX A6000 (48 GB each), using a temperature setting of 0.7 for sampling.

\subsection {Evaluating One-to-Many Anonymization Quality}
We evaluate the anonymization fidelity of the proposed CON-QA framework in comparison to the hide model from the HaS \cite{chen2023hide} framework, focusing specifically on complex, multi-PII contractual queries drawn from the CUAD-QA dataset. This comparison is motivated by the necessity to assess how effectively each framework safeguards sensitive information while maintaining semantic integrity in realistic legal QA settings.

The hide model, fine-tuned on a fixed dataset, frequently reuses surrogate entities across queries and fails to anonymize all detected PII instances. Such reuse introduces potential linkability and privacy leakage, particularly in enterprise contexts where repeated exposure of semantically similar placeholders may compromise confidentiality.

In contrast, CON-QA employs a structured anonymization approach based on precise entity-level detection, followed by query-specific one-to-many surrogate set generation. These surrogate mappings are created independently for each session using in-house Qwen prompts, ensuring: (i) complete entity coverage, (ii) high surrogate diversity, and (iii) zero cross-query reuse.
To evaluate these anonymization strategies, both models are tested on 50 contractual queries and chunks, each containing multiple types of sensitive entities with an average of 18 PII entities. We compute and compare the following anonymization quality metrics:
\begin{itemize}

 \item Coverage (↑): The proportion of all detected sensitive entities that are successfully replaced with anonymized surrogates during the anonymization phase. Higher values indicate more comprehensive protection.

 \item Surrogate Reuse (↓) : The percentage of surrogate entities that appear more than once across anonymized query-document pairs. Lower reuse values correspond to better surrogate diversity and reduced pattern leakage.

 \item Unique Surrogates (↑): The average number of distinct anonymized replacements used per sensitive entity across the dataset. Higher values indicate greater diversity and resilience against entity inference.

 \item Linkability (↓): A qualitative measure of the extent to which surrogate entities can be correlated across multiple queries or sessions. Lower linkability implies stronger privacy guarantees through minimized pattern repetition.

 \item Missed Entities (↓): The number of sensitive entities identified during NER that were not replaced during anonymization. Fewer missed entities suggest more complete anonymization coverage and stronger privacy protection.
\end{itemize}

\begin{table}
\centering
\caption{Comparison of Hide Model vs. Con-QA Metrics}
\label{tab:qa_anonymization_comparison}
\begin{tabular}{lccccc}
\toprule
\textbf{Framework} & \textbf{Cov. (\%)} & \textbf{Reuse (\%)} & \textbf{Uniq. Sur(\%).} & \textbf{Link.(\%)} & \textbf{Missed(\%)} \\
\midrule
Hide Model HaS        & 76.1.5  & 46.71  & 59.12   & 46.71 & 31.38 \\
\textbf{Con-QA}    & \textbf{99.11} & \textbf{1.34} & \textbf{98.76} & \textbf{1.23} & \textbf{0.89} \\
\bottomrule
\end{tabular}
\end{table}

The anonymization performance comparison between the proposed CON-QA framework and the Hide model from the HaS framework \cite{chen2023hide} Table~\ref{tab:qa_anonymization_comparison} reveals CON-QA's significant advantage in enterprise-grade contractual QA settings. CON-QA achieves a high coverage of 99.11\%, indicating near-complete replacement of sensitive entities, while the Hide model lags at 76.15\%, often missing critical PII elements. Moreover, CON-QA maintains a surrogate reuse rate of only 1.34\%, compared to 46.71\% in Hide, thereby drastically reducing linkability risk (1.23\% vs. 46.71\%) and enhancing resistance against surrogate pattern inference. The framework also demonstrates superior surrogate diversity with 98.76\% unique surrogates, outperforming Hide's 59.12\%, and leaves only 0.89\% of entities unanonymized, whereas Hide misses 31.38\%. These metrics collectively underscore CON-QA’s effectiveness in preserving privacy without compromising semantic integrity, making it a robust solution for real-world legal document QA where accurate yet confidential information retrieval is critical.

\subsection{Evaluating Deanonymization Quality: many-to-one mapping}
We evaluate the deanonymization quality of the proposed CON-QA framework against the Seek model from the Hide-and-Seek (HaS) framework, focusing on the accurate recovery of original entities in contractual QA settings. This comparison is motivated by the need to assess entity fidelity and semantic precision after anonymized queries and documents are processed by cloud-based LLMs. The Seek model relies on a learned decoder trained to infer original entities from anonymized outputs, often suffering from incomplete restoration or hallucinated entities due to model generalization errors. In contrast, CON-QA employs a deterministic, session-specific many-to-one mapping strategy, which ensures that each anonymized entity is accurately and unambiguously reversed using many-to-one mapping. This eliminates recovery ambiguity and enhances interpretability. We assess both methods across 50 CUAD-QA and We quantify recovery performance using two metrics: BERTScore and Entity Accuracy, as evaluation metrics.
\begin{table}[h]
\centering
\caption{Deanonymization Quality Comparison between Con-QA and Seek Model}
\label{tab:deanonymization_quality}
\begin{tabular}{lcc}
\toprule
\textbf{Model} & \textbf{Entity Accuracy (\%)} ↑ & \textbf{BERTScore (\%)} ↑ \\
\midrule
Seek Model     & 89.00 & 81.10 \\
\textbf{Con-QA (Ours)} & \textbf{98.00} & \textbf{83.00} \\
\bottomrule
\end{tabular}
\end{table}

\paragraph{Analysis}:We evaluate the deanonymization performance of the Con-QA framework against the Seek model from the HaS framework, focusing on accurate restoration of sensitive entities in contractual QA. As shown in Table~\ref{tab:deanonymization_quality}, Con-QA achieves superior entity accuracy (98.00\%) compared to the Seek model (89.00\%), demonstrating reliable and unambiguous recovery through its deterministic, session-specific many-to-one mapping strategy. While Con-QA also outperforms Seek on BERTScore (83.00\% vs. 81.10\%), we note that BERTScore slightly underrepresents the semantic fidelity of recovered answers. This is primarily due to the metric's sensitivity to surface-level lexical overlap with short reference answers, whereas CON-QA's responses are often longer and more contextually enriched. To address this limitation and validate the answer quality, we additionally conduct expert human evaluation with contract domain expert, as described in the following subsection.

\subsection{Expert Human Evaluation}
Given the semantic complexity and privacy sensitivity of contractual QA, we conducted a qualitative expert evaluation to assess the fidelity of the CON-QA framework. While automated metrics like BERTScore showed marginal dips due to strict surface-level comparisons with concise reference answers, our recovered answers often provided more fine-grained and explanatory responses—crucial in legal contexts. To address this, two expert groups were involved: (i) eight NLP domain Data scientist with privacy and language understanding backgrounds rated 30 QA sets QA on five parameters using a 5-point scale; (ii) two legal domain experts from academia and industry evaluated the same sets using four domain-specific criteria. This dual-layer assessment highlights both the privacy strength and domain-aligned utility of CON-QA in enterprise contract QA. The evaluations yielded an overall accuracy of 85.83\% from NLP experts and 91.66\% from legal experts, confirming that despite minor drops in summarization-type queries, CON-QA maintains high answer quality and outperforms surface-based metrics in real-world.showing graph~\ref{fig:humanEval}.The detailed evaluation graphs for both expert criteria and parameter-based ratings are provided in the project GitHub repository\footnote{\url{https://github.com/IOTPL-research/con_QA/tree/main/result/nlp_domain_participant_rating}}
along with corresponding query-wise results to ensure transparency and clarity.

\begin{figure*}[t]
\centering
\includegraphics[width=16cm, height = 8cm]{ 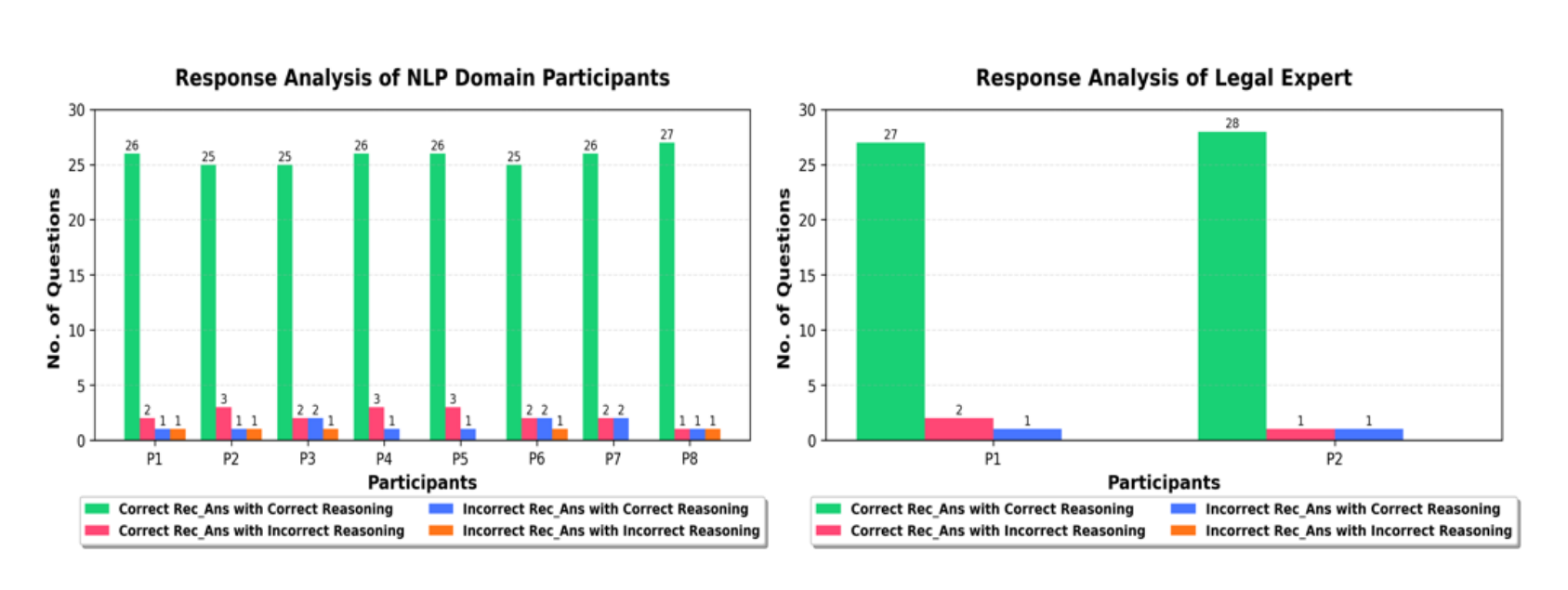}

\caption{NLP domain participant response (left) and  Contract Expert (right) }

\label{fig:humanEval}
\end{figure*}
.
\section{Conclusion}
We introduced CON-QA, a hybrid privacy-preserving framework that seamlessly integrates local and cloud-based LLMs to enable secure question answering over enterprise contracts. Through structured anonymization and session-consistent deanonymization, CON-QA ensures strong protection of sensitive contractual entities while preserving semantic fidelity and answer quality. Experimental evaluation has been done additionally we also conducted expert human evaluations to validate its effectiveness in maintaining legal accuracy and privacy, demonstrating its practical relevance for secure, real-world enterprise applications.


\bibliography{acml25}






\end{document}